\pdfoutput=1
\documentclass[11pt,a4paper]{article}
\usepackage[hyperref]{emnlp2020}
\usepackage{times}
\usepackage{latexsym}

\usepackage{sec/package}
\usepackage{microtype}

\aclfinalcopy 
\def\aclpaperid{733} 

\title{
Tasty Burgers, Soggy Fries: Probing Aspect Robustness in \\ Aspect-Based Sentiment Analysis
}

\author{Xiaoyu Xing\textsuperscript{\rm 1}$^{*}$ {} {} Zhijing Jin\textsuperscript{\rm 2}\thanks{{} {} Equal Contribution. Please email (or cc') both first authors for correspondence.} {} {} {} Di Jin\textsuperscript{\rm 3} {} {} Bingning Wang\textsuperscript{\rm 4} {} {} Qi Zhang\textsuperscript{\rm 1} {} {} Xuanjing Huang\textsuperscript{\rm 1}\\  
\textsuperscript{\rm 1}Fudan University
\textsuperscript{\rm 2}Max Planck Institute,
\textsuperscript{\rm 3}CSAIL, MIT,
\textsuperscript{\rm 4}Sogou Inc.
\\
\texttt{\{xyxing18,qz,xjhuang\}@fudan.edu.cn}
\\
\texttt{zjin@tuebingen.mpg.de}
\\
\texttt{
jindi15@mit.edu,
wangbingning@sogou-inc.com
} 
}  

\date{}

\begin{document}
\maketitle

\begin{abstract}
    Aspect-based sentiment analysis (ABSA) aims to predict the sentiment towards a specific aspect in the text. However, existing ABSA test sets cannot be used to probe whether a model can distinguish the sentiment of the target aspect from the non-target aspects.
    To solve this problem, we develop a simple but effective approach to enrich ABSA test sets. Specifically, we generate new examples to disentangle the confounding sentiments of the non-target aspects from the target aspect's sentiment. Based on the SemEval 2014 dataset, we construct the Aspect Robustness Test Set (ARTS) as a comprehensive probe of the aspect robustness of ABSA models. Over 92\% data of ARTS show high fluency and desired sentiment on all aspects by human evaluation. Using ARTS, we analyze the robustness of nine ABSA models, and observe, surprisingly, that their accuracy drops by up to 69.73\%. We explore several ways to improve aspect robustness, and find that adversarial training can improve models' performance on ARTS by up to 32.85\%.\footnote{Our code and new test set are available at \url{https://github.com/zhijing-jin/ARTS_TestSet}.
    }
\end{abstract}

\section{Introduction}\label{sec:intro}

Aspect-based sentiment analysis (ABSA) is an advanced sentiment analysis task that aims to classify the sentiment towards a specific aspect (e.g., \textit{burgers} or \textit{fries} in the review ``Tasty \textit{burgers}, and crispy \textit{fries}.''). The key to a strong ABSA model is it is sensitive to \textit{only} the sentiment words of the target aspect, and therefore not be interfered by the sentiment of any non-target aspect. Although state-of-the-art models have shown high accuracy on existing test sets, we still question their robustness. Specifically, given the \textit{prerequisite} that a model outputs correct sentiment polarity for the test example, we have the following questions:

\begin{enumerate}[itemsep=-1ex,topsep=1ex,label=(Q\arabic*)]
    \item If we reverse the sentiment polarity of the target aspect, can the model change its prediction accordingly?
    \label{item:intro_ex0}
    \item If the sentiments of all non-target aspects become opposite to the target one, can the model still make the correct prediction?
    \label{item:intro_ex1}
    \item If we add more non-target aspects with sentiments opposite to the target one, can the model still make the correct prediction?
    \label{item:intro_ex2}
\end{enumerate}

A robust ABSA model should both meet the prerequisite and have affirmative answers to all the questions above. For example, if a model makes the correct sentiment classification (i.e., positive) for \textit{burgers} in the original sentence ``Tasty \textit{burgers}, and crispy fries'', it should flip its prediction (to negative) when seeing the new context ``Terrible \textit{burgers}, but crispy fries''. Hence, these questions together form a probe to verify if an ABSA model has high \textbf{\textit{aspect robustness}}. 

\begin{table*}[t]
  \centering
  \resizebox{\textwidth}{!}{
    \begin{tabular}{p{0.5cm} p{8cm}p{7cm}}
    \toprule
    \multicolumn{1}{c}{\textbf{SubQ.}} 
    & \multicolumn{1}{c}{\textbf{Generation Strategy}} &\multicolumn{1}{c}{\textbf{Example}}
    \\ \hline
    \textbf{Prereq.} & \textbf{\textsc{Source}}: The original sample from the test set & Tasty \textbf{burgers}, and crispy fries. (Tgt: burgers)
    \\
    \textbf{Q1} & \textbf{\textsc{RevTgt}}: Reverse the sentiment of the 
    \textit{target} aspect  & \ul{Terrible} \textbf{burgers}, but crispy fries. 
    \\
    \textbf{Q2} & \textbf{\textsc{RevNon}}: Reverse the sentiment of the \textit{non-target} aspects with originally the same sentiment as target & Tasty \textbf{burgers}, but \ul{soggy} fries. 
    \\
    \textbf{Q3} & \textbf{\textsc{AddDiff}}: Add aspects with the \textit{opposite} sentiment from the target aspect & Tasty \textbf{burgers}, crispy fries\ul{, but poorest service ever!}
    \\
    \bottomrule
    \end{tabular}%
  }
  \caption{The generation strategies and examples of the prerequisite (Prereq) and three questions \ref{item:intro_ex0}-\ref{item:intro_ex2}. Each example are annotated with the \textbf{target aspect (Tgt)}, and \ul{altered sentence parts}.
  }
  \label{tab:intro_4strategies}
\end{table*}
Unfortunately, existing ABSA datasets have very limited capability to probe the \textit{aspect robustness}. For example, the Twitter dataset \cite{dong-etal-2014-adaptive} has only one aspect per sentence, so the model does not need to discriminate against non-target aspects. In the most widely used SemEval 2014 Laptop and Restaurant datasets~\cite{pontiki-etal-2014-semeval}, for 83.9\% and 79.6\% instances in the test sets, the sentiments of the target aspect, and all non-target aspects are all the same. Hence, we cannot decide whether models that make correct classifications attend only to the target aspect, because they may also wrongly look at the non-target aspects, which are \textit{confounding factors}. Only a small portion of the test set can be used to answer our target questions proposed in the beginning. Moreover, when we test on the subset of the test set (59 instances in Laptop, and 122 instances in Restaurant) where the target aspect sentiment differs from all non-target aspect sentiments (so that the confounding factor is disentangled), the best model \cite{xu2019bert} drops from 78.53\% to 59.32\% on Laptop and from 86.70\% to 63.93\% on Restaurant. This implies that the success of previous models may over-rely on the confounding non-target aspects, but not necessarily on the target aspect only. However, \textit{no} datasets can be used to analyze the aspect robustness more in depth.


We develop an automatic generation framework that takes as input the original test instances from SemEval 2014, and applies three generation strategies showed in Table~\ref{tab:intro_4strategies}. New test instances generated by \textsc{RevTgt}, \textsc{RevNon}, and \textsc{AddDiff} can be used to answer the questions \ref{item:intro_ex0}-\ref{item:intro_ex2}, respectively. The generated new instances largely overlap with the content and aspect terms of the original instances, but manage to disentangle the confounding sentiment polarity of non-target aspects from the target, as showed in the examples in Table~\ref{tab:intro_4strategies}.
In this way, we produce an ``all-rounded'' test set
that can test whether a model robustly captures the target sentiment instead of other irrelevant clues.

We enriched the laptop dataset by 294\% from 638 to 1,877 instances and the restaurant dataset by 315\% from 1,120 to 3,530 instances.
By human evaluation, more than 92\% of the new aspect robustness test set (ARTS) shows high fluency, and desired sentiment on all aspects. Our ARTS test set is in line with other recent works on NLP challenge sets \cite{mcCoy2019right,matt2020evaluating}.
Using our new test set,  we analyze the aspect robustness of nine existing models. Experiment results show that their performance degrades by up to 55.64\% on Laptop and 69.73\% on Restaurant. We also use our generation strategy to conduct adversarial training and find it improves aspect robustness by at least 11.87\% and at most 35.37\% across various models.

The contributions of our paper are as follows:
\begin{enumerate}[itemsep=-1ex,topsep=0.5ex]
    \item We develop simple but effective automatic generation methods that generate new test instances (with over 92\% accuracy by human evaluation) to challenge the aspect robustness. 
    \item We construct ARTS, a new test set targeting at aspect robustness for ABSA models, and propose a new metric, Aspect Robustness Score.
    \item We probe the aspect robustness of nine models, and reveal up to 69.73\% performance drop on ARTS compared with the original test set.
    \item We provide several solutions to enhance aspect robustness for ABSA models, including adversarial training detailed in Section~\ref{sec:adv_training}.
\end{enumerate}

\section{Data Generation}\label{sec:method}

As shown in Table~\ref{tab:intro_4strategies}, we aim to build a systematic method to generate all possible aspect-related alternations, in order to remove the confounding factors in the existing ABSA data. In the following, we will introduce three different ways to disentangle the sentiment of the target aspect from sentiments of non-target aspects.

\subsection{\textsc{RevTgt}}
The first strategy is to generate sentences that reverse the original sentiment of the target aspect. The word spans of each aspect's sentiment of SemEval 2014 data are provided by \cite{FanWDHC19}. We design two methods to reverse the sentiment, and one additional step of conjunction adjustment on top of the two methods to polish the resulting sentence. 

\begin{table}[ht]
\small
  \centering
    \begin{tabular}{p{1.5cm} p{5.5cm}}
    \toprule
     \multicolumn{1}{c}{\textbf{Strategy}} & \multicolumn{1}{c}{\textbf{Example}}  \\
    \midrule
    \multirow{2}{*}{Flip Opinion} & It's \textbf{light} and \textbf{easy} to \underline{transport}.
    \\ 
     & $\rightarrow$ It's \textbf{heavy} and \textbf{difficult} to \underline{transport}. 
    \\ \hline
    \multirow{2}{*}{Add Negation} & The \underline{menu} \textbf{changes} seasonally.
    \\ 
     & $\rightarrow$ The \underline{menu} \textbf{does not change} seasonally. 
    \\ \hline
    Adjust & The food is good, \textbf{and} the \underline{decor} is \textbf{nice}.
    \\ 
     Conjunctions & $\rightarrow$ The food is good, \textbf{but} the \underline{decor} is \textbf{nasty}.
    \\
    \bottomrule
    \end{tabular}%
  \caption{Three strategies and examples of \textsc{RevTgt}.}
  \label{tab:revtgt}
\end{table}

\begin{table*}[ht]
\small
  \centering
    \begin{tabular}{p{4.5cm} p{2cm} p{2cm} p{2cm} p{2cm}}
    \toprule
     \multicolumn{1}{c}{\textbf{Strategy}} & \multicolumn{4}{c}{\textbf{Example}}  \\
    \midrule
    \multirow{2}{*}{\textbf{Original sentence \& sentiment}} & \multicolumn{4}{l}{It has great \underline{food} and a reasonable price, but the service is poor.}
    \\ 
     &  \textbf{\possentiment{(Tgt) \underline{food}:$+$}}  &  \textbf{\possentiment{price:$+$}}  & \textbf{\negsentiment{service:$-$}}  & \textbf{\neusentiment{overall:$\Circle$}} 
    \\
    \textbf{\textsc{RevNon}}
    \\
    \quad Flip same-sentiment non-target aspects (and adjust conjunctions) & \multicolumn{4}{l}{It has great \underline{food} \textbf{but} an \textbf{unreasonable} price, \textbf{and} the service is poor. }
    \\ 
    \quad Exaggerate opposite-sentiment  & \multicolumn{4}{l}{It has great \underline{food} \textbf{but} an \textbf{unreasonable} price, \textbf{and} the service is \textbf{extremely} poor. }
    \\
    non-target aspects &  \textbf{\possentiment{(Tgt) \underline{food}:$+$}}  &  \textbf{\negsentiment{price:$-$}}  & \textbf{\negsentiment{service:$--$}}  & \textbf{\negsentiment{overall:$-$}} 
    \\ 
    \bottomrule
    \end{tabular}%
  \caption{The generation process of \textsc{RevNon}. The \ul{target aspect} (Tgt), and sentiments of all aspects are annotated.}
  \label{tab:revnon}
\end{table*}

\paragraph{Flip Opinion Words} Suppose we have the sentence ``\textit{Tasty} \textbf{burgers} and crispy fries,'' where the sentiment term for the target aspect is \textit{Tasty}. We aim to generate a new sentence that flips the sentiment \textit{Tasty}. A baseline approach is antonym replacement by looking up WordNet \cite{miller95wordnet}. However, due to polysemy, the simple lookup is very likely to derive an inappropriate antonym and cause incompatibility with the context. Among the retrieved set of antonyms, we only keep words with the same Part-of-Speech (POS) tag as original, using the stanza package\footnote{\url{https://stanfordnlp.github.io/stanza/}} which takes the context into account by the state-of-the-art neural network-based model.\footnote{For the candidate filter, we do not use GPT-2 perplexity because its low accuracy, e.g., 38.4\% on a random sample set. And its output is also less interprettable than the POS filter.} Lastly, in the case of multiple antonyms, we prioritize the words that are already in the existing vocabulary, and then randomly select an antonym from the candidate set.

\paragraph{Add Negation} As the above strategy of flipping by the antonym is constrained by whether appropriate antonyms are available. For those cases without suitable antonyms, including long phrases, we add negation according to the linguistic features. In most cases, the sentiment expression is an adjective or verb term, so we simply add negation (i.e., ``not'') in front of it. If the sentiment term is not an adjective or verb, we add negation to its closest verb. For example, in Table~\ref{tab:revtgt}, there are no available antonyms for ``change'' in the original example ``The \underline{menu} \textbf{changes} seasonally.'', so we simply negate it as ``The \underline{menu} \textbf{does not change} seasonally.''

\paragraph{Adjust Conjunctions} 
As pinpointed in Section~\ref{sec:intro}, 79.6$\sim$83.9\% of the original test data of SemEval 2014 \cite{pontiki-etal-2014-semeval} have the same sentiment for all aspects. A possible result of reverting one aspect's sentiment is that the other aspects' sentiments will be opposite to the altered one. So we need to adjust the conjunctions for language fluency.
If the two closest surrounding sentiments of a conjunction word have the same polarity, then cumulative conjunctions such as ``and'' should be applied; otherwise, we should adopt adversative conjunctions such as ``but.''
In the example in Table~\ref{tab:revtgt}, after flipping the sentiment, we derive the example ``The food is good, \textbf{and} the decor is nasty'' which is very unnatural, so we replace the conjunction ``and'' with ``but,'' and thus generate the example ``The food is good, \textbf{but} the decor is nasty.'' 

\subsection{\textsc{RevNon}}
Changing the target sentiment by \textsc{RevTgt} can test if a model is sensitive enough towards the target-aspect sentiment, but we need to further complement this probe by perturbing the sentiments of the non-target aspects (\textsc{RevNon}). As showed in Table~\ref{tab:revnon}, for all the non-target aspects with the same sentiment as the target aspect's, we reverse their sentiments using the same method as \textsc{RevTgt}. And for all the remaining non-target aspects, whose sentiments are already opposite from the target sentiment, we exaggerate the extent by randomly adding an adverb (e.g., ``very'', ``really'' and ``extremely'') from a dictionary of adverbs of degree that is collected based on the training set. The resulting test example will be a solid proof of the ABSA quality, because only the target aspect has the desired sentiment, and all non-target aspects have been flipped to or exaggerated with the opposite sentiment.

\subsection{\textsc{AddDiff}}
The first two strategies, \textsc{RevTgt} and \textsc{RevNon}, have explored how the sentiment changes of existing aspects will challenge an ABSA model, and \textsc{AddDiff} further investigate if adding more non-target aspects can confuse the model. Moreover, the existing SemEval 2014 test sets have only on average 2 aspects per sentence, but the real-world applications can have more aspects. With these motivations, we develop \textsc{AddDiff} as follows.

We first form a set of aspect expressions $\mathrm{AspectSet}$\footnote{The full $\mathrm{AspectSet}$ is available on our GitHub.} by extracting all aspect expressions from the entire dataset. 
Specifically, for each example in the dataset, we first identify each sentiment term (e.g., ``reasonable'' in ``Food at a reasonable price'') and then extract its linguistic branch as the aspect expression (e.g., ``at a reasonable price'') by pretrained constituency parsing \cite{joshi2018extending}. Table~\ref{tab:adddiff} shows several examples of $\mathrm{AspectSet}$ in the restaurant domain.

\begin{table}[ht]

  \centering
  \resizebox{0.9\columnwidth}{!}{
    \begin{tabular}{p{1.5cm} p{5.5cm}}
    \toprule
     \multicolumn{1}{c}{\textbf{Sentiment}} & \multicolumn{1}{c}{\textbf{Aspect Expression}}  \\
    \midrule
    \multirow{3}{*}{Positive} & staff is \textit{friendly} and \textit{knowledgeable}
    \\ 
     & desserts are \textit{out of this world}
     \\
     & texture is a \textit{velvety}
    \\ \hline
    \multirow{3}{*}{Negative} & service is \textit{severely slow}
    \\ 
    & dining experience is \textit{miserable}
    \\
    & tables are \textit{uncomfortably close}
    \\
    \bottomrule
    \end{tabular}%
  }
  \caption{Example aspect expressions from $\mathrm{AspectSet}$ of the restaurant domain.}
  \label{tab:adddiff}
\end{table}

Using the $\mathrm{AspectSet}$, we randomly sample 1-3 aspects that are not mentioned in the original test sample and whose sentiments are different from the target aspect's, and then append these to the end of the original example. For example, ``Great food and best of all GREAT beer!'' $\xrightarrow{\textsc{AddDiff}}$ ``Great food and best of all GREAT beer, \textit{but management is less than accommodating, music is too heavy, and service is severely slow}.''
In this way, \textsc{AddDiff} enables the advanced testing of whether the model will be confused when there are more irrelevant aspects with opposite sentiments.

\section{ARTS Dataset}
\subsection{Overview}
Our source data is the most\footnote{We surveyed deep learning-based ABSA papers from 2015 to 2020 at top conferences (ACL, EMNLP, NAACL, NeurIPS, ICLR, ICML, AAAI, IJCAI). Among the 63 ABSA papers, 50 use SemEval 2014 Laptop and Restaurant, which is the top 1 widely used dataset.} widely used ABSA dataset, SemEval 2014 Laptop and Restaurant Reviews \cite{pontiki-etal-2014-semeval}.\footnote{\url{http://alt.qcri.org/semeval2014/task4/}}
We follow \cite{wang2016attention,MaLZW17,xu2019bert} to remove instances with conflicting polarity and only keep positive, negative, and neutral labels. We use the train-dev split as in \cite{xu2019bert}. The resulting Laptop dataset has 2,163 training, 150 validation, and 638 test instances, and Restaurant has 3,452 training, 150 validation, and 1,120 test instances.

Building upon the original SemEval 2014 data, we generate enriched test sets of 1,877 instances (294\% of the original size) in the laptop domain, and 3,530 instances (315\%) in the restaurant domain using generation method introduced in Section~\ref{sec:method}. The statistics of our ARTS test set are in Table~\ref{tab:dataset_overview}. (A more detailed explanations of the number of instances generated by each strategy is in Appendix~\ref{appd:gen_stats}.)
\begin{table}[ht]
  \centering
  \resizebox{0.9\columnwidth}{!}{
    \begin{tabular}[\columnwidth]{lll}
    \toprule
    & \multicolumn{1}{c}{\textbf{Laptop}} & \multicolumn{1}{c}{\textbf{Restaurant}}  \\
    \midrule
    Original Test Set & 638 & 1,120
    \\ 
    Enriched Test Set & 1,877 & 
    3,530 \\
    Relative Size & 294.20\% &315.17\%
    \\ \hline
    \multicolumn{3}{c}{\textbf{\textit{Fluency Check}}}
    \\
    Accepted Instances &1,732  & 3,260
    \\ 
    Fixed Instances &145  &270 
    \\
    Acceptance Rate & 92.27\% &92.35 \%
    \\
    Inter-Agreement & 91.10\% & 92.69\%
    \\ \hline
    \multicolumn{3}{c}{\textbf{\textit{Sentiment Check}}}
    \\
    Accepted Instances &1,763  &3,362
    \\ 
    Fixed Instances &114  &168 
    \\
    Acceptance Rate & 93.93\% & 95.24\%
    \\
    Inter-Agreement & 94.14\% & 95.61\%
    \\
    \bottomrule
    \end{tabular}%
  }
  \caption{Overall statistics of the ARTS test set and results of fluency and sentiment checks.}
  \label{tab:dataset_overview}
\end{table}
\subsection{Quality Inspection}
We conduct human evaluation to validate the generation quality of our ARTS dataset on two criteria:
\begin{enumerate}[itemsep=-1ex,topsep=0ex]
    \item \textbf{Fluency}: Does the generated sentence maintain the fluency of the source sentence?
    \item \textbf{Sentiment Correctness}: Does the sentiment of each aspect have the desired polarity?
\end{enumerate}
\begin{itemize}[itemsep=-1ex,topsep=-1ex]
    \item \textsc{RevTgt}: Is the target sentiment reversed?
\end{itemize}
\begin{itemize}[nolistsep]    
    \item \textsc{RevNon}: For non-target aspects with originally the same sentiment as the target, is it reversed? For the rest, are they exaggerated?
    \item \textsc{AddDiff}: Is the target sentiment unchanged?
\end{itemize}

Each task is completed by two native-speaker judges. We first calculate the inter-agreement rate of the human annotators, and then resolve the divergent opinions on samples that they disagree with. We accept the samples that both judges considered as correct or are resolved to be correct after our check. Finally, we ask the annotators to fix the rejected samples by minimal edit which does not change the aspect term or the sentence meaning, but satisfies both criteria.
\paragraph{Fluency Check}
The evaluation results on fluency are showed in Table~\ref{tab:dataset_overview}. Most samples (92.27\% of  Laptop and 92.35\% of Restaurant test sets) are accepted as fluent text. The inter-agreement rate between the two human judges is also high, 91.10\% and 92.69\% on the two datasets.
\paragraph{Sentiment Check}
We also evaluate the sentiment correctness of the generated text. Note that for \textsc{RevNon}, we count the samples with all ``yes'' answers as accepted samples. Overall, the acceptance rate of the generated samples is 93.93\% on Laptop and 95.24\% on Restaurant, along with inter-agreement rates of over 94.14\% on both datasets.

\subsection{Dataset Analysis}
After checking the quality of our enriched ARTS test set, we analyze the dataset characteristics and make comparisons with the original test sets.
\begin{table}[ht]
\centering
\resizebox{\columnwidth}{!}{
    \begin{tabular}{lllll}
    \toprule
    & \multicolumn{2}{c}{\textbf{Laptop}} & \multicolumn{2}{c}{\textbf{Restaurant}}
    \\
    & Ori & ARTS & Ori & ARTS
    \\
    \midrule
    \#Words/Sent & 18.56 & 22.27 & 19.37 & 23.15
    \\
    Vocab Size & 1565 & 1746 & 2197 & 2451
    \\
    \multicolumn{3}{l}{Labels}
    \\
    \quad Positive & 341 & 883 & 728 & 1953
    \\
    \quad Negative & 128 & 587 & 196 & 1104
    \\
    \quad Neutral & 169 & 407 & 196 & 473
    \\
    \quad \#Positive/\#Negative & 2.66 & 1.5 & 3.71 & 1.77
    \\
    \multicolumn{3}{l}{Aspect-Related Challenge}
    \\
    \quad \#Aspects/Sent & 2.05 & 2.75 & 2.57 & 3.28
    \\
    \quad Opp. Nontgt $\geq$ 1  & 16\% & 59\% & 20\% & 67\%
    \\
    \quad Opp. Nontgt $=$ All  & 9\% & 38\% & 11\% & 42\%
    \\
    \quad \#Opp. Nontgt/Sent & 0.23 & 1.16 & 0.27 & 1.39
    \\
    \bottomrule
    \end{tabular}%
}
\caption{Characteristics of the ARTS test sets in comparison to the Original (``Ori'') Laptop and Restaurant test sets.}
  \label{tab:dataset_char}
\end{table}

For \textit{general statistics}, we can see from Table~\ref{tab:dataset_char} that the sentence length in the new test set is on average 4 words more than the original, and the vocabulary size is also larger by around two hundred. \textit{For the label distribution}, we can see that the new test set has an increasing number of all labels, and especially balances the ratio of positive-to-negative labels from the original 2.66 to 1.5 on Laptop, and from 3.71 to 1.77 on Restaurant. 

For the \textit{aspect-related challenge} in the test set, the new test set, first of all, has a larger number of aspects per sentence than the original. Our test set also features the higher disentanglement of the target aspect from the non-target aspects that have the same sentiment as the target: the portion of samples with at least one non-target aspects of sentiments different from the target is 59$\sim$67\%, and on average 45\% higher than the original test sets. And the portion of the most challenging samples where all non-target aspects have sentiments different from the target one on the new test set is on average 30\% more than that of the original test set. The average number of non-target aspects with opposite sentiments per sample in the new test set is on average 5 times that of the original set.

\subsection{Aspect Robustness Score (ARS)}
As mentioned in Section~\ref{sec:intro}, a model is considered to have high aspect robustness
if it satisfies both the prerequisite and all three questions \ref{item:intro_ex0}-\ref{item:intro_ex2}. 
So we propose a novel metric, Aspect Robustness Score (ARS), that counts the correct classification of the source example and all its variations (\textsc{RevTgt}, \textsc{RevNon}, and \textsc{AddDiff}) as one unit of correctness. Then we apply the standard calculation of accuracy. Note that the three variations correspond to questions \ref{item:intro_ex0}-\ref{item:intro_ex2}, respectively. 

\section{Evaluating ABSA Models}\label{sec:experiments}

We use our enriched test set as a comprehensive test on the aspect robustness of ABSA models.

\subsection{Models}
For a comprehensive overview of the ABSA field, we conduct extensive experiments on models with a variety of neural network architectures.  

\textbf{TD-LSTM:} \cite{tang2015effective} uses two Long Short-Term Memory Networks (LSTM) to encode the preceding and following contexts of the target aspect (inclusive) and concatenate the last hidden states of the two LSTMs to make the sentiment classification.

\textbf{AttLSTM:} \citet{wang2016attention} apply an Attention-based LSTM on the concatenatation of the aspect and word embeddings of each token.

\textbf{GatedCNN:} \citet{xue2018aspect} use a Gated Convolutional Neural Networks (CNN) that applies a Tanh-ReLU gating mechanism to the CNN-encoded text with aspect embeddings.

\textbf{MemNet:} \citet{tang2016aspect} use memory networks to store the sentence as external memory and calculate the attention with the target aspect.

\textbf{GCN:} Aspect-specific Graph Convolutional Networks (GCN) \cite{zhang2019aspect} first applies GCN over the syntax tree of the sentence and then imposes an aspect-specific masking layer on its top.

\textbf{BERT:} \citet{xu2019bert} uses a BERT-based baseline \cite{devlin2018bert} and takes as input the concatenation of the aspect term and the sentence.

\textbf{BERT-PT:} \citet{xu2019bert} post-train BERT on other review datasets such as Amazon laptop reviews \cite{he2016ups} and Yelp Dataset Challenge reviews, and finetune on ABSA tasks.

\textbf{CapsBERT:} \cite{jiang2019challenge} encode the sentence and the aspect term with BERT, and then feed it into Capsule Networks to predict the polarity.

\textbf{BERT-Sent:} For more in-depth analysis, we also implement a sentence classification baseline. BERT-Sent takes as input sentences without aspect information, and predicts the ``global'' sentiment. We use it because if other ABSA models fails to pay attention to aspects, they will degenerate to a sentence classifier. If so, they will resemble BERT-Sent, which performs well on original tests and badly on ARTS. So BERT-Sent is a reference to check degenerated aspect-level models.

\subsection{Implementation Details}

For all existing models, we use the authors' official implementation. For our self-proposed BERT-Sent, we use Adam optimizer with
a learning rate of 5e-5, weight decay of 0.01, 
and batch size of 32. We apply the $l_2$  regularization with $\lambda=10^{-4}$, and train 50 epochs.
Note that the tokenization of the ASTS dataset is the same as the original SemEval 2014, as we prepared the new test set by inverting the NLTK tokenization rules we used when applying the generation strategies.\footnote{We used the metanl package to detokenize: \url{https://github.com/commonsense/metanl}.}


\subsection{Results on ARTS}

We list the accuracy\footnote{For ABSA, accuracy is the standard metric to be reported \cite{wang2016attention, xue2018aspect, tang2016aspect}.} of the nine models on the 
\begin{table*}[!t]
\centering
\small
\resizebox{\textwidth}{!}{
\begin{tabular}{lllll}
\toprule
Model & \multicolumn{1}{c}{Entire Test}
& \multicolumn{1}{c}{\textsc{RevTgt} Subset}
& \multicolumn{1}{c}{\textsc{RevNon} Subset}
& \multicolumn{1}{c}{\textsc{AddDiff} Subset}
\\
& \multicolumn{1}{c}{Ori $\rightarrow$ New (Change)}
& \multicolumn{1}{c}{Ori $\rightarrow$ New (Change)}
& \multicolumn{1}{c}{Ori $\rightarrow$ New (Change)}
& \multicolumn{1}{c}{Ori $\rightarrow$ New (Change)}
\\
\hline
\multicolumn{5}{l}{\textbf{\textit{Laptop Dataset}}}
\\
MemNet & 64.42 $\rightarrow$ 16.93 (\textcolor[rgb]{0.76,0,0}{$\downarrow$47.49})$^\star$ & 72.10 $\rightarrow$ 28.33 (\textcolor[rgb]{0.72,0,0}{$\downarrow$43.77})$^\star$ & 82.22 $\rightarrow$ 79.26 (\textcolor[rgb]{0.28,0,0}{$\downarrow$02.96}) & 64.42 $\rightarrow$ 56.58 (\textcolor[rgb]{0.33,0,0}{$\downarrow$07.84})$^\star$
\\
GatedCNN & 65.67 $\rightarrow$ 10.34 (\textcolor[rgb]{0.85,0,0}{$\downarrow$55.33})$^\star$ & 75.11 $\rightarrow$ 24.03 (\textcolor[rgb]{0.8,0,0}{$\downarrow$51.08})$^\star$ & 83.70 $\rightarrow$ 78.52 (\textcolor[rgb]{0.31,0,0}{$\downarrow$05.18}) & 65.67 $\rightarrow$ 45.14 (\textcolor[rgb]{0.47,0,0}{$\downarrow$20.53})$^\star$
\\
AttLSTM & 67.55 $\rightarrow$ 09.87 (\textcolor[rgb]{0.87,0,0}{$\downarrow$57.68})$^\star$ & 72.96 $\rightarrow$ 27.04 (\textcolor[rgb]{0.74,0,0}{$\downarrow$45.92})$^\star$ & 85.93 $\rightarrow$ 75.56 (\textcolor[rgb]{0.36,0,0}{$\downarrow$10.37})$^\star$ & 67.55 $\rightarrow$ 39.66 (\textcolor[rgb]{0.55,0,0}{$\downarrow$27.89})$^\star$
\\
TD-LSTM & 68.03 $\rightarrow$ 22.57 (\textcolor[rgb]{0.74,0,0}{$\downarrow$45.46})$^\star$ & 73.39 $\rightarrow$ 29.83 (\textcolor[rgb]{0.72,0,0}{$\downarrow$43.56})$^\star$ & 83.70 $\rightarrow$ 77.04 (\textcolor[rgb]{0.32,0,0}{$\downarrow$06.66}) & 68.03 $\rightarrow$ 60.66 (\textcolor[rgb]{0.33,0,0}{$\downarrow$07.37})$^\star$
\\
GCN & 72.41 $\rightarrow$ 19.91 (\textcolor[rgb]{0.81,0,0}{$\downarrow$52.50})$^\star$ & 78.33 $\rightarrow$ 35.62 (\textcolor[rgb]{0.71,0,0}{$\downarrow$42.71})$^\star$ & 88.89 $\rightarrow$ 74.81 (\textcolor[rgb]{0.4,0,0}{$\downarrow$14.08})$^\star$ & 72.41 $\rightarrow$ 52.51 (\textcolor[rgb]{0.46,0,0}{$\downarrow$19.90})$^\star$
\\
BERT-Sent & 73.04 $\rightarrow$ 17.40 (\textcolor[rgb]{0.85,0,0}{$\downarrow$55.64})$^\star$ & 78.76 $\rightarrow$ 59.44 (\textcolor[rgb]{0.46,0,0}{$\downarrow$19.32})$^\star$ & 88.15 $\rightarrow$ 42.22 (\textcolor[rgb]{0.74,0,0}{$\downarrow$45.93})$^\star$ & 73.04 $\rightarrow$ 34.64 (\textcolor[rgb]{0.66,0,0}{$\downarrow$38.40})$^\star$
\\
CapsBERT & 77.12 $\rightarrow$ 25.86\footnote{CapsBERT hand-crafted the Capsule Guided Routing specifically for the restaurant domain, so it fails significantly.} (\textcolor[rgb]{0.8,0,0}{$\downarrow$51.26})$^\star$ & 80.69 $\rightarrow$ 57.73 (\textcolor[rgb]{0.5,0,0}{$\downarrow$22.96})$^\star$ & 88.89 $\rightarrow$ 49.63 (\textcolor[rgb]{0.67,0,0}{$\downarrow$39.26})$^\star$ & 77.12 $\rightarrow$ 45.14 (\textcolor[rgb]{0.59,0,0}{$\downarrow$31.98})$^\star$
\\
BERT & 77.59 $\rightarrow$ 50.94 (\textcolor[rgb]{0.54,0,0}{$\downarrow$26.65})$^\star$ & 83.05 $\rightarrow$ 65.02 (\textcolor[rgb]{0.44,0,0}{$\downarrow$18.03})$^\star$ & 93.33 $\rightarrow$ 71.85 (\textcolor[rgb]{0.48,0,0}{$\downarrow$21.48})$^\star$ & 77.59 $\rightarrow$ 71.00 (\textcolor[rgb]{0.32,0,0}{$\downarrow$06.59})$^\star$
\\
BERT-PT & 78.53 $\rightarrow$ 53.29 (\textcolor[rgb]{0.52,0,0}{$\downarrow$25.24})$^\star$ & 82.40 $\rightarrow$ 60.09 (\textcolor[rgb]{0.49,0,0}{$\downarrow$22.31})$^\star$ & 93.33 $\rightarrow$ 83.70 (\textcolor[rgb]{0.35,0,0}{$\downarrow$09.63})$^\star$ & 78.53 $\rightarrow$ 75.71 (\textcolor[rgb]{0.28,0,0}{$\downarrow$02.82})
\\
\textbf{Average} & 71.60 $\rightarrow$ 25.23 (\textcolor[rgb]{0.75,0,0}{$\downarrow$46.37})$^\star$ & 77.42 $\rightarrow$ 43.01 (\textcolor[rgb]{0.62,0,0}{$\downarrow$34.41})$^\star$ & 87.57 $\rightarrow$ 70.29 (\textcolor[rgb]{0.44,0,0}{$\downarrow$17.28})$^\star$ & 71.60 $\rightarrow$ 53.45 (\textcolor[rgb]{0.45,0,0}{$\downarrow$18.15})$^\star$
\\ \hline

\multicolumn{5}{l}{\textbf{\textit{Restaurant Dataset}}}
\\
MemNet & 75.18 $\rightarrow$ 21.52 (\textcolor[rgb]{0.83,0,0}{$\downarrow$53.66})$^\star$ & 80.73 $\rightarrow$ 27.54 (\textcolor[rgb]{0.82,0,0}{$\downarrow$53.19})$^\star$ & 84.46 $\rightarrow$ 73.65 (\textcolor[rgb]{0.37,0,0}{$\downarrow$10.81})$^\star$ & 75.18 $\rightarrow$ 60.71 (\textcolor[rgb]{0.41,0,0}{$\downarrow$14.47})$^\star$
\\
GatedCNN & 76.96 $\rightarrow$ 13.12 (\textcolor[rgb]{0.94,0,0}{$\downarrow$63.84})$^\star$ & 85.11 $\rightarrow$ 23.17 (\textcolor[rgb]{0.92,0,0}{$\downarrow$61.94})$^\star$ & 88.06 $\rightarrow$ 72.97 (\textcolor[rgb]{0.41,0,0}{$\downarrow$15.09})$^\star$ & 76.96 $\rightarrow$ 54.91 (\textcolor[rgb]{0.49,0,0}{$\downarrow$22.05})$^\star$
\\
AttLSTM & 75.98 $\rightarrow$ 14.64 (\textcolor[rgb]{0.91,0,0}{$\downarrow$61.34})$^\star$ & 82.98 $\rightarrow$ 28.96 (\textcolor[rgb]{0.83,0,0}{$\downarrow$54.02})$^\star$ & 86.26 $\rightarrow$ 61.26 (\textcolor[rgb]{0.52,0,0}{$\downarrow$25.00})$^\star$ & 75.98 $\rightarrow$ 52.32 (\textcolor[rgb]{0.5,0,0}{$\downarrow$23.66})$^\star$
\\
TD-LSTM & 78.12 $\rightarrow$ 30.18 (\textcolor[rgb]{0.77,0,0}{$\downarrow$47.94})$^\star$ & 85.34 $\rightarrow$ 34.99 (\textcolor[rgb]{0.79,0,0}{$\downarrow$50.35})$^\star$ & 88.51 $\rightarrow$ 75.68 (\textcolor[rgb]{0.39,0,0}{$\downarrow$12.83})$^\star$ & 78.12 $\rightarrow$ 70.18 (\textcolor[rgb]{0.34,0,0}{$\downarrow$07.94})$^\star$
\\
GCN & 77.86 $\rightarrow$ 24.73 (\textcolor[rgb]{0.82,0,0}{$\downarrow$53.13})$^\star$ & 86.76 $\rightarrow$ 35.58 (\textcolor[rgb]{0.8,0,0}{$\downarrow$51.18})$^\star$ & 88.51 $\rightarrow$ 79.50 (\textcolor[rgb]{0.35,0,0}{$\downarrow$09.01})$^\star$ & 77.86 $\rightarrow$ 65.00 (\textcolor[rgb]{0.39,0,0}{$\downarrow$12.86})$^\star$
\\
BERT-Sent & 80.62 $\rightarrow$ 10.89 (\textcolor[rgb]{1.0,0,0}{$\downarrow$69.73})$^\star$ & 89.60 $\rightarrow$ 44.80 (\textcolor[rgb]{0.73,0,0}{$\downarrow$44.80})$^\star$ & 89.86 $\rightarrow$ 57.21 (\textcolor[rgb]{0.6,0,0}{$\downarrow$32.65})$^\star$ & 80.62 $\rightarrow$ 30.89 (\textcolor[rgb]{0.78,0,0}{$\downarrow$49.73})$^\star$
\\
CapsBERT & 83.48 $\rightarrow$ 55.36 (\textcolor[rgb]{0.55,0,0}{$\downarrow$28.12})$^\star$ & 89.48 $\rightarrow$ 71.87 (\textcolor[rgb]{0.44,0,0}{$\downarrow$17.61})$^\star$ & 90.99 $\rightarrow$ 74.55 (\textcolor[rgb]{0.43,0,0}{$\downarrow$16.44})$^\star$ & 83.48 $\rightarrow$ 77.86 (\textcolor[rgb]{0.31,0,0}{$\downarrow$05.62})$^\star$
\\
BERT & 83.04 $\rightarrow$ 54.82 (\textcolor[rgb]{0.55,0,0}{$\downarrow$28.22})$^\star$ & 90.07 $\rightarrow$ 63.00 (\textcolor[rgb]{0.54,0,0}{$\downarrow$27.07})$^\star$ & 91.44 $\rightarrow$ 83.33 (\textcolor[rgb]{0.34,0,0}{$\downarrow$08.11})$^\star$ & 83.04 $\rightarrow$ 79.20 (\textcolor[rgb]{0.29,0,0}{$\downarrow$03.84})$^\star$
\\
BERT-PT & 86.70 $\rightarrow$ 59.29 (\textcolor[rgb]{0.54,0,0}{$\downarrow$27.41})$^\star$ & 92.20 $\rightarrow$ 72.81 (\textcolor[rgb]{0.46,0,0}{$\downarrow$19.39})$^\star$ & 92.57 $\rightarrow$ 81.76 (\textcolor[rgb]{0.37,0,0}{$\downarrow$10.81})$^\star$ & 86.70 $\rightarrow$ 80.27 (\textcolor[rgb]{0.32,0,0}{$\downarrow$06.43})$^\star$
\\
\textbf{Average} & 79.77 $\rightarrow$ 31.62 (\textcolor[rgb]{0.77,0,0}{$\downarrow$48.15})$^\star$ & 86.92 $\rightarrow$ 44.75 (\textcolor[rgb]{0.7,0,0}{$\downarrow$42.17})$^\star$ & 88.96 $\rightarrow$ 73.32 (\textcolor[rgb]{0.42,0,0}{$\downarrow$15.64})$^\star$ & 79.77 $\rightarrow$ 63.48 (\textcolor[rgb]{0.43,0,0}{$\downarrow$16.29})$^\star$
\\
\bottomrule
\end{tabular}
}
\caption{Model accuracy on Laptop and Restaurant data. We compare the accuracy on the \textbf{Ori}ginal and our \textbf{New} test sets (\textit{Ori $\rightarrow$ New}), and calculate the \textit{change} of accuracy. Besides the Entire Test Set, we also list accuracy on subsets where the generation strategies \textsc{RevTgt}, \textsc{RevNon} and \textsc{AddDiff} can be applied. The accuracy of \textit{Entire Test-New} is calculated using ARS.
$\star$ indicates whether the performance drop is statistically significant (with p-value $\leq 0.05$ by Welch's $t$-test).}
\label{tab:main_res}
\end{table*}
Laptop and Restaurant test sets in Table \ref{tab:main_res}. 
For \textit{Entire Test-New} in Table~\ref{tab:main_res}, accuracy is calculated using ARS. Supplementary to ARS, Table \ref{tab:main_res} also decomposes ARS into single-strategy scores (the right three columns) by splitting the entire ARTS test set into three subsets according to the corresponding data generation techniques. Each of the single-strategy scores explains from one perspective the reason for large performance drop in ARS, which will be elaborated later.

\paragraph{Overall Performance} 
On the entire test set, we can see that the accuracy of all models on the original test set is very high, achieving up to 78.53\% on Laptop and 86.70\% on Restaurant, but it drops drastically ($\downarrow$69\%$\sim\downarrow$25\%) on our new test sets.

\paragraph{Performance of Different Models}

From the overall performance on our new test set, we can see that BERT models on average are more robust to the aspect-targeted challenges that our new test set poses. The most effective model BERT-PT scores the best on both original accuracy and robustness. It has 53.29\% ARS on Laptop and 59.29\% on Restaurant. However, the accuracy of non-BERT models  on average drops drastically to under 30\% by over $\downarrow$50\%.

\paragraph{Performance on Different Subsets}
We list in detail the performance of each model on the three subsets of our new test set: \textsc{RevTgt}, \textsc{RevNon}, and \textsc{AddDiff}. 
They correspond to the three questions \ref{item:intro_ex0}-\ref{item:intro_ex2}.
\textsc{RevTgt} on average induces the most performance drop, as it requires the model to pay precise attention to the target sentiment words. \textsc{RevNon} makes the performance of the sentence classifier BERT-Sent drops the most by up to $\downarrow$45.93\%, and the model CapsBERT also drops by up to $\downarrow$39.26\%. The last subset \textsc{AddDiff} causes most non-BERT models to drop significantly, indicating that these models are not robust enough against an increased number of non-target aspects, which should have been irrelevant.

\paragraph{Laptop vs. Restaurant}
The performance drop on Restaurant is higher than that on Laptop. There are two possible reasons: (1) the original performance on restaurant is higher, and (2) the new test set is more challenging in the Restaurant domain.
We verify this by calculating the relative drop ($\frac{\text{new}-\text{old}}{\text{old}}$) in addition to the reported absolute values of the change. The relative drop on Laptop is 64.76\%, which is higher than 60.36\% on Restaurant.
For the laptop dataset, both the lower original performance and the larger relative decrease of performance might be due to the nature of the dataset. For example, Laptop restaurant has far fewer training data than Restaurant, which makes the models less accurate originally and weaker on ARTS. 

\section{Analysis}\label{sec:analysis}
\subsection{Variations of Generation Strategies}

\paragraph{Combining Multiple Strategies}
Each sample in the ARTS test set is generated by one of the three strategies. However, it is also worth exploring whether combining several strategies can make a more challenging probe on the aspect robustness of ABSA models. As a case study, we analyze the model robustness against test samples generated by the combination of \textsc{RevNon+AddDiff}.
\begin{table}[!ht]
\centering
\small
\resizebox{\columnwidth}{!}{
\begin{tabular}{lll}
\toprule
Model & \multicolumn{1}{c}{Laptop} & \multicolumn{1}{c}{Restaurant}
\\
& \multicolumn{1}{c}{Ori $\rightarrow$ New (Change)} 
& \multicolumn{1}{c}{Ori $\rightarrow$ New (Change)}
\\
\hline 
MemNet & 82.22 $\rightarrow$ 72.59 ($\downarrow$09.63) & 84.46 $\rightarrow$ 50.90 ($\downarrow$33.56)$^\star$
\\
GatedCNN & 84.44 $\rightarrow$ 59.26 ($\downarrow$25.18)$^\star$ & 87.84 $\rightarrow$ 53.83 ($\downarrow$34.01)$^\star$
\\
AttLSTM & 85.93 $\rightarrow$ 51.85 ($\downarrow$34.08)$^\star$ & 86.26 $\rightarrow$ 38.06 ($\downarrow$48.20)$^\star$
\\
TD-LSTM & 83.70 $\rightarrow$ 68.89 ($\downarrow$14.81)$^\star$ & 88.51 $\rightarrow$ 65.99 ($\downarrow$22.52)$^\star$
\\
GCN & 88.89 $\rightarrow$ 60.74 ($\downarrow$28.15)$^\star$ & 88.51 $\rightarrow$ 72.52 ($\downarrow$15.99)$^\star$
\\
BERT-Sent & 88.15 $\rightarrow$ 11.85 ($\downarrow$76.30)$^\star$ & 89.86 $\rightarrow$ 11.94 ($\downarrow$77.92)$^\star$
\\
CapsBERT & 90.37 $\rightarrow$ 24.44 ($\downarrow$65.93)$^\star$ & 90.99 $\rightarrow$ 66.89 ($\downarrow$24.10)$^\star$
\\
BERT & 93.33 $\rightarrow$ 68.15 ($\downarrow$25.18)$^\star$ & 91.44 $\rightarrow$ 76.58 ($\downarrow$14.86)$^\star$
\\
BERT-PT & 93.33 $\rightarrow$ 78.52 ($\downarrow$14.81)$^\star$ & 92.57 $\rightarrow$ 78.60 ($\downarrow$13.97)$^\star$
\\
\textbf{Average} &87.57 $\rightarrow$ 55.14 ($\downarrow$32.43) $^\star$ &88.96 $\rightarrow$ 57.26 ($\downarrow$31.70)$^\star$
\\
\bottomrule
\end{tabular}
}
\caption{The accuracy of each model on the original test set and the new test set generated by \textsc{RevNon+AddDiff} in laptop and restaurant domains.}
\label{tab:revnon_and_adddiff}
\end{table}
By comparing the performance decrease caused by \textsc{RevNon+AddDiff} in Table~\ref{tab:revnon_and_adddiff} and by only \textsc{RevNon} and \textsc{AddDiff} in Table~\ref{tab:main_res}, we can see that the accuracy of each model decreases by a much larger extent on \textsc{RevNon+AddDiff} than either of \textsc{RevNon} and \textsc{AddDiff}.
The performance drop by the \textsc{RevNon+AddDiff} subset is almost the sum of \textsc{RevNon} and \textsc{AddDiff} in most cases, and sometimes larger than the sum, for example, in the case of BERT-PT on Laptop where the sum of the performance drop by the two separate strategies 9.63\%+2.82\% is smaller than the combined strategy's 14.81\%.

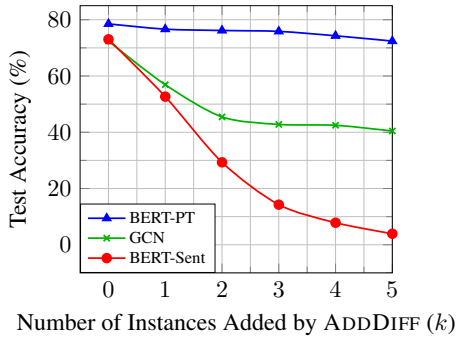
\begin{figure}[ht!]
  \centering
   \resizebox{0.8\columnwidth}{!}{%
            \begin{tikzpicture}
        \pgfplotsset{
            scale only axis,
            xmin=-0.5, xmax=5,
            xtick={0,...,5},
            legend style={at={(0,0)},anchor=south west},
        }
        \begin{axis}[
          grid=both,
          legend style={cells={align=left},nodes={scale=0.7, transform shape},
          legend cell align={left}},
          ymin=-10, ymax=85,
          xlabel=Number of Instances Added by \textsc{AddDiff} ($k$),
          ylabel=Test Accuracy (\%),
          minor tick num=1,
          every axis plot/.append style={thick}
        ]
        \addplot[smooth,mark=triangle*,blue]
          coordinates{
            (0, 78.53)
            (1, 76.65)
            (2, 76.18)
            (3, 75.86)
            (4, 74.29)
            (5, 72.41)
        }; \addlegendentry{BERT-PT}
        
        \addplot[smooth,mark=x,black!30!green]
          coordinates{
            (0, 72.41)
            (1, 56.90)
            (2, 45.45)
            (3, 42.79)
            (4, 42.48)
            (5, 40.44)
        }; \addlegendentry{GCN}
        
        \addplot[smooth,mark=*,red]
          coordinates{
            (0, 73.04)
            (1, 52.66)
            (2, 29.31)
            (3, 14.26)
            (4, 7.84)
            (5, 3.92)
        };
        \addlegendentry{BERT-Sent}
        
        \end{axis}
        \end{tikzpicture}
        }
\caption{Accuracy of BERT-PT, GCN, and BERT-Sent on the test samples in the laptop domain generated by \textsc{AddDiff}$(k)$ where $k$ varies from 1 to 5.}
\label{fig:add_diff_plot}
\end{figure}

\paragraph{\textsc{AddDiff} with More Aspects}
Some strategies such as \textsc{AddDiff} can be parameterized by $k$, where $k$ is the number of additional non-target aspects to be added. We select three models (the best, the worst, and an average-performing one), and plot their accuracy on test samples generated by \textsc{AddDiff}$(k)$ on Laptop in Figure~\ref{fig:add_diff_plot}. As $k$ gets larger, the test samples become more difficult. The sentence classification baseline BERT-Sent drops drastically, BERT-PT remains high, and GCN lies in the middle.
\subsection{How to Effectively Model the Aspect?}\label{sec:discuss_models}
An important usage of our ARTS is to understand what model components are key to aspect robustness. We list the aspect-specific mechanisms of all models according to the ascending order of their ARS on Laptop dataset in Table~\ref{tab:model_overview}. We can see that for BERT-based models, BERT-PT, which is further trained on large review corpora, gets the best accuracy and aspect robustness. More complicated structures like CapsBERT underperforms the basic BERT by 25.08\%.
\begin{table}[ht!]

  \centering
  \resizebox{\columnwidth}{!}{%
    \begin{tabular}{lcccc}
    \toprule
     \multicolumn{1}{c}{\textbf{Model}} & \textbf{ARS} & \textbf{Asp+W Emb} & \textbf{Posi-Aware} & \textbf{Asp Att}
     \\
    \hline
\textbf{AttLSTM} & 9.87 & \cmark & \xmark & \cmark
\\
\textbf{GatedCNN} & 10.34 & \cmark & \xmark & \cmark
\\
\textbf{MemNet} & 16.93 & \xmark & \xmark & \cmark
\\
\textbf{GCN} & 19.91 & \xmark & \possentiment{\cmark} & \cmark
\\
\textbf{TD-LSTM} & 22.57 & \xmark & \possentiment{\cmark} & \xmark
\\
\textbf{CapsBERT} & 25.86 & \xmark & \xmark & \cmark
\\
\textbf{BERT} & 50.94 & \xmark & \xmark & \xmark
\\
\textbf{BERT-PT} & 53.29 & \xmark & \xmark & \xmark
\\
    \bottomrule
    \end{tabular}%
  }
  \caption{Models in the ascending order of their ARS on Laptop. We list their aspect-specific mechanisms, including concatenating the aspect and word embeddings (Asp+W Emb), position-aware mechanism for aspects (Posi-Aware), and attention using the aspect (Asp Att). We highlight \possentiment{\cmark} for Posi-Aware as it is the most related to aspect robustness for non-BERT models.}
  \label{tab:model_overview}
\end{table}

Among the non-BERT models, the aspect position-aware models TD-LSTM and GCN are the most robust, as they have a stronger sense of the location of the target aspect in a sentence. On the contrary, 
the other models with poorer robustness (9.87\%$\sim$16.93\% in Table~\ref{tab:model_overview}) only use mechanisms such as aspect-based attention, or concatenating the aspect embedding to the word embedding.

To summarize, the main takeaways are
\begin{itemize}[nolistsep] 
    \item For BERT models, additional pretraining is the most effective.
    \item For non-BERT models, explicit position-aware designs lead to more aspect robustness.
\end{itemize}
\begin{table*}[t!]
    \centering
    \begin{subtable}[c]{0.3\textwidth}
    \small
        \begin{tabular}{lcc}
        \toprule
        \multicolumn{1}{c}{\multirow{2}{*}{Model}} & \multicolumn{2}{c}{MAMS} \\
        \multicolumn{1}{c}{} &
          O$\rightarrow$O &
          O$\rightarrow$N \\ \hline
        MemNet & 70.51 & 37.80\\
        GatedCNN & 66.02 & 32.93\\
        AttLSTM & 67.14 & 39.67 \\
        TD-LSTM & 77.62 & 49.25 \\
        GCN & 76.95 &  47.98 \\
        BERT-Sent & 49.25 & 10.48 \\
        CapsBERT & 83.38 &  60.18 \\
        BERT & 84.51 &  61.38 \\
        BERT-PT & 85.10 &  64.37 \\
        \bottomrule
       \end{tabular}
      \caption{Accuracy of each model trained on the MAMS \textbf{O}riginal training data and evaluated on the \textbf{O}riginal test data (O$\rightarrow$O), as well as the \textbf{N}ew test set generated by our models (O$\rightarrow$N).}
       \label{tab:mams}
    \end{subtable}
    \hfill
    \begin{subtable}[h]{0.65\textwidth}
    \small
        \begin{tabular}{lccc|ccc}
        \toprule
        \multicolumn{4}{c|}{Restaurant} &
        \multicolumn{3}{c}{Laptop} \\ 
          \multicolumn{1}{c}{O$\rightarrow$O} &
          O$\rightarrow$N &
          MAMS$\rightarrow$N &
          Adv$\rightarrow$N &
          O$\rightarrow$O &
          O$\rightarrow$N &
          Adv$\rightarrow$N 
        \\ \hline
        75.18 & 21.52 & 24.02 & 37.95 & 64.42 & 16.93 & 31.82\\
        76.96 & 13.13 & 18.48 & 37.50 & 65.67 & 10.34 & 41.85\\
        75.98 & 14.64 & 22.32 & 48.66 & 67.55 & 9.87 & 42.63 \\
        78.12 & 30.18 & 41.60 & 62.76 & 68.03 & 22.57 & 54.86\\
        77.86 & 24.73 & 46.51 & 61.52 & 72.41 & 19.91 & 56.43\\
        80.62 & 10.89 & 12.95 & 45.80 & 73.04 & 17.40 & 53.92\\
        83.66 & 55.36 & 61.43 & 75.80 & 76.80 & 25.86 & 61.23\\
        83.04 & 54.82 & 62.77 & 74.82 & 77.59 & 50.94 & 65.67\\
        86.70 & 59.29 & 62.77 & 74.64 & 78.53 & 53.29 & 66.93\\
        \bottomrule
        \end{tabular}
        \caption{Accuracy of each model trained on the \textbf{O}riginal data and evaluated on the \textbf{O}riginal test set (O$\rightarrow$O), and the \textbf{N}ew test set (O$\rightarrow$N), as well as that trained on the \textbf{Adv}ersarial data and evaluated on the \textbf{N}ew test set (Adv$\rightarrow$N).
        For Restaurant, we also test models trained on \textbf{MAMS} dataset and tested on the \textbf{N}ew test set of Restaurant (MAMS$\rightarrow$N).}
        \label{tab:data_augmentation}
     \end{subtable}
     \caption{Improvements on the new test set using different training data.}
\label{tab:more_training}
\end{table*}
\subsection{Does a More Diverse Training Set Help?} \label{sec:better_training}

A recent dataset, Multi-Aspect Multi-Sentiment (MAMS) \cite{jiang2019challenge}, is collected from the same data source as the SemEval 2014 Restaurant dataset~\cite{ganu2009beyond}. However, its sentences are more complicated, each having at least two aspects with different sentiment polarities.

We use this dataset to inspect whether a stronger training set can help improve aspect robustness.

\paragraph{Training and Testing on MAMS}

Table~\ref{tab:mams} checks the aspect robustness of models trained on MAMS using the original MAMS test set (O$\rightarrow$O) and the new test set that we produced by applying the same generation strategies to its test set (O$\rightarrow$N). Models trained and tested on MAMS have a smaller decrease rate than those on the Restaurant dataset. This shows that a more challenging training set can make models more robust.

\paragraph{Training on MAMS and Testing on Restaurant}
As MAMS and Restaurant are collected from the same source data, we test whether MAMS-trained models perform well on the new test set of Restaurant (in the column ``MAMS$\rightarrow$N'' of Table~\ref{tab:data_augmentation}). We can see that all models trained on MAMS are more robust than those trained on the Restaurant dataset. For example, the accuracy of BERT and BERT-PT on the new test set is lifted up to 62.77\%. 

\subsection{Does Adversarial Training Help?} \label{sec:adv_training}
Although the MAMS described in Section~\ref{sec:better_training} provides a training set with diversity, it remains difficult to improve aspect robustness for other domains, or future new datasets. Therefore, we propose a flexible method, adversarial training, for aspect robustness, which is applicable to any given dataset.

We conducted adversarial training on the Laptop and Restaurant datasets, and analyze its effect in Table~\ref{tab:data_augmentation}. Specifically, for the column ``Adv$\rightarrow$N'', we generated an additional training set by applying the three proposed strategies on training data, then trained models on the augmented data obtained by combining the original training data and this newly generated data, and finally evaluated on the ARTS test data. This practice follows Table 7 of \cite{zhang2019paws} which is a similar stream of work as ours for the paraphrasing domain.

In both Restaurant and Laptop domains, adversarial training (Adv$\rightarrow$N) leads to significant performance improvement than only training on the original datasets (O$\rightarrow$N). On the Restaurant datasets, adversarial training is even more effective than training on MAMS, because our generated data instances comprehensively covered all possible perturbations of the non-target aspects, and naturally collected datasets might not be comparable.

\section{Error Analysis for Data Generation}
We analyze the error types in the subset of ARTS that was fixed by human judges. Two most significant error types are wrong antonyms ($\sim$2\%), such as ``the weight of the laptop is \possentiment{light}$\rightarrow$\negsentiment{dark}'', and negation which causes grammatical errors ($\sim$1.1\%). In future work, we can fix the latter by applying a grammatical error correction system on top of our generation. Also, \textsc{RevTgt} and \textsc{RevNon} cannot be applied to 1.4$\sim$6.6\% instances with complicated sentiment expressions which rely on commonsense. For example, ``a 2-hour wait'' is negative bust too difficult to alter in our current generation framework. It needs more advanced models such as text style transfer~\cite{shen2017style,jin2019imat}.

\section{Related Work}
\paragraph{Robustness in NLP}
Robustness in NLP has attracted extensive attention in recent works \cite{HsiehCJWHH19, LiMJ16a}. As a popular method to probe the robustness of models, adversarial text generation becomes an emerging research field in NLP. Techniques include adding extraneous text to the input \cite{jia-liang-2016-data}, character-level noise \cite{belinkov2018synthetic,ebrahimi2018hotflip}, and word replacement \cite{alzantot2018generating,jin2019bert}. Using the adversarial generation techniques, new adversarial test sets are proposed for several tasks such as paraphrasing \cite{zhang2019paws} and entailment \cite{glockner2018breaking,mcCoy2019right}. 

\paragraph{Aspect-Based Sentiment Analysis}
ABSA has emerged as an active research area recently. Early works hand-craft sentiment lexicons and syntactic features for rule-based classifiers \cite{jiang2011target, kiritchenko2014nrc}. Recent neural network-based models use architectures such as LSTM \cite{tang2015effective}, CNN \cite{xue2018aspect}, Attention mechanisms \cite{wang2016attention}, Capsule Network \cite{jiang2019challenge}, and the pretrained model BERT \cite{xu2019bert}. Similar to the motivation in our paper, some work shows preliminary speculation that the current ABSA datasets might be downgraded to sentence-level sentiment classification \cite{xu2019failure}.

\section{Conclusion}

In this paper, we proposed a simple but effective mechanism to generate test instances to probe the aspect robustness of the models. We enhanced the original SemEval 2014 test sets by 294\% and 315\% in laptop and restaurant domains. Using our new test set, we probed the aspect robustness of nine ABSA models, and discussed model designs and better training that can improve the robustness.

\section*{Acknowledgments}
We appreciate Professor Rada Mihalcea for her insights that helped us plan this research, Pengfei Liu for valuable suggestions on writing, and Yuchun Dai for helping to code some functions in our annotation tool. We also want to convey special thanks to Mahi Shafiullah and Osmond Wang for brilliant suggestions on the wording of the title.

\bibliography{anthology,emnlp2020}
\bibliographystyle{acl_natbib}

\appendix

\section{Appendices}
\label{sec:appendix}
\subsection{Test Set Generation Details}\label{appd:gen_stats}

In our main paper, we mentioned that the size of our enriched ARTS test set is 294\% of the original Laptop data size, and 315\% of the original Restaurant test data size. These two ratios should ideally be both 400\%, because there are three generation strategies, plus one original sentence. However, this gap is because not every original test sentence can qualify for every generation strategy. The number of instances generated by each strategy is shown in Table \ref{tab:statistics}.

\begin{table}[h]
\centering
\resizebox{\columnwidth}{!}{%
\begin{tabular}{lcccc}
\toprule
\multicolumn{1}{c}{Dataset} & Ori Test 
& \textsc{RevTgt} & \textsc{RevNon} & \textsc{AddDiff} \\ \midrule
Laptop     & 638  
& 466  & 135  & 638  \\ 
Restaurant & 1120 
& 846  & 444  &1120 \\ 
MAMS       & 1336 
& 789  & 402  & 1336 \\
\bottomrule
\end{tabular}%
}
\caption{The statistics of dataset. }
\label{tab:statistics}
\end{table}
Although AddDiff can apply to all test cases, the other two strategies, \textsc{RevTgt} and \textsc{RevNon}, cannot apply to all instances. 
For \textsc{RevTgt} and \textsc{RevNon}, we need to flip opinion words, so we can only apply these two strategies on instances with explicit opinion words. The number of opinion words is the main bottleneck for \textsc{RevTgt}. Specifically, \citet{fan2019target} provide opinion words for 466 instances of Laptop, and 846 instances of Restaurant. Since \textsc{RevTgt} is applicable to these instances, the number of new test instances in ARTS generated by \textsc{RevTgt} are the same as the number of opinion word-available instances.

In addition to the opinion word constraint, \textsc{RevNon} has some further requirements. We filter out instances which have only one aspect (224 in Laptop, and 263 in Restaurant), as well as instances where the opinion words of the target aspect are overlapped with the opinion words of the non-target aspect (102 in Laptop and 132 in Restaurant). Also we did not consider instances of which all the sentiment of non-target aspects are neutral (4 in Laptop, and 7 in Restaurant).

\end{document}